\documentclass[letterpaper, 10 pt, journal, twoside]{IEEEtran}
\usepackage{booktabs}
\usepackage{fancyhdr}
\usepackage{graphics}
\usepackage{graphicx}
\usepackage{epsfig}
\usepackage{color}
\usepackage[dvipsnames]{xcolor}
\usepackage[format=plain,
            labelfont=it,
            labelsep=period]{caption}
\usepackage{subcaption}
\usepackage{wrapfig}
\usepackage{float}
\usepackage{cite}
\usepackage{times}
\usepackage{amsmath}
\usepackage{amssymb}
\usepackage{amsfonts}
\usepackage{mathtools}
\usepackage{url}
\usepackage{algorithm}
\usepackage{algpseudocode}
\definecolor{citecolor}{HTML}{0071bc}
\usepackage[pagebackref=true,breaklinks=true,colorlinks,linkcolor=BrickRed,citecolor=citecolor,urlcolor=citecolor,bookmarks=false]{hyperref}

\definecolor{citecolor}{HTML}{0071bc}

\newcommand\blfootnote[1]{%
  \begingroup
  \renewcommand\thefootnote{}\footnote{#1}%
  \addtocounter{footnote}{-1}%
  \endgroup
}

\begin{document}
%
\title{Look Closer: Bridging Egocentric and Third-Person Views with Transformers for Robotic Manipulation}
%
%
%


\author{
Rishabh Jangir\textsuperscript{*1}\quad
Nicklas Hansen\textsuperscript{*1}\quad
Sambaran Ghosal\textsuperscript{1}\quad
Mohit Jain\textsuperscript{1}\quad
Xiaolong Wang\textsuperscript{1}

}
%
%

\markboth{IEEE Robotics and Automation Letters. Preprint Version. Accepted January, 2022}
{Jangir \MakeLowercase{\textit{et al.}}: Look Closer: Bridging Egocentric and Third-Person Views with Transformers for Robotic Manipulation} 

%




\twocolumn[{%
\renewcommand\twocolumn[1][]{#1}%
\maketitle
\vspace{-0.2in}
\begin{center}
    \centering
    \captionsetup{type=figure}
    \includegraphics[width=0.92\textwidth]{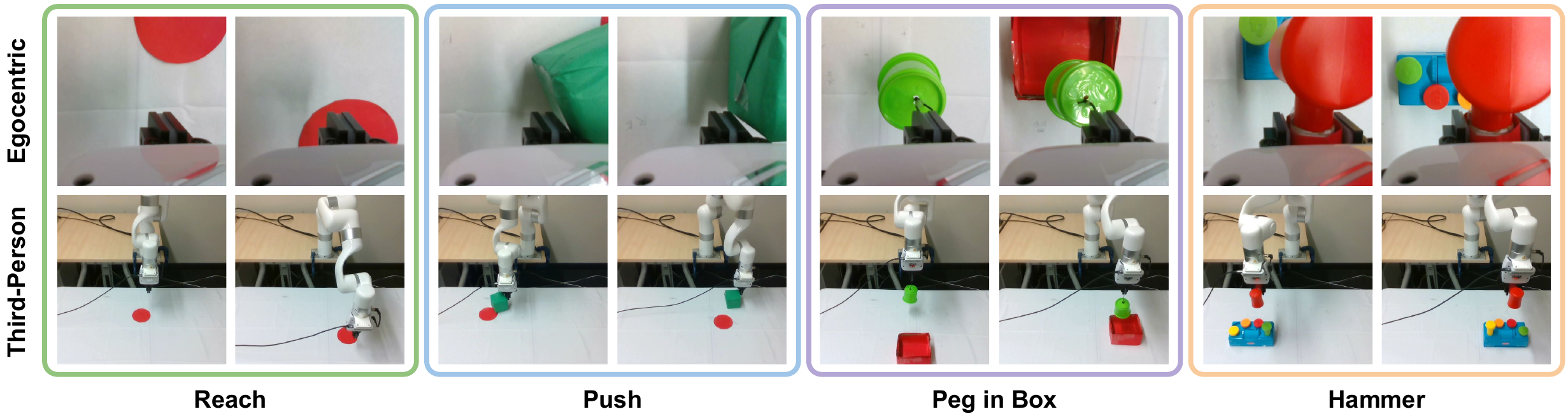}
    \captionof{figure}{\textbf{Multi-view robotic manipulation.} We propose a method for vision-based robotic manipulation that fuses egocentric and third-person views using a cross-view attention mechanism. Our method learns a policy using reinforcement learning that successfully transfers from simulation to a real robot, and solves precision-based manipulation tasks directly from uncalibrated cameras, without access to state information, and with a high degree of variability in task configurations.}
    \label{fig:teaser}
\end{center}%
}]

\begin{abstract}
Learning to solve precision-based manipulation tasks from visual feedback using Reinforcement Learning (RL) could drastically reduce the engineering efforts required by traditional robot systems. However, performing fine-grained motor control from visual inputs alone is challenging, especially with a static third-person camera as often used in previous work. We propose a setting for robotic manipulation in which the agent receives visual feedback from both a third-person camera and an egocentric camera mounted on the robot's wrist. While the third-person camera is static, the egocentric camera enables the robot to actively control its vision to aid in precise manipulation. To fuse visual information from both cameras effectively, we additionally propose to use Transformers with a \textit{cross-view} attention mechanism that models spatial attention from one view to another (and vice-versa), and use the learned features as input to an RL policy. Our method improves learning over strong single-view and multi-view baselines, and successfully transfers to a set of challenging manipulation tasks on a real robot with uncalibrated cameras, no access to state information, and a high degree of task variability. In a hammer manipulation task, our method succeeds in $75\%$ of trials versus $38\%$ and $13\%$ for multi-view and single-view baselines, respectively. Project website can be found  \href{https://jangirrishabh.github.io/lookcloser/}{https://jangirrishabh.github.io/lookcloser/}.

\blfootnote{
Manuscript received: September, 09, 2021; Revised December, 09, 2021; Accepted December, 30, 2021. \\
\indent *Equal contribution. All the authors are with \textsuperscript{1}University of California San Diego, CA, USA. Correspondence at {\tt\footnotesize xiw012@ucsd.edu}.\\ 
\indent This paper was recommended for publication by Editor Tamim Asfour upon evaluation of the Associate Editor and Reviewers' comments. \\
\indent This work was supported by grants from NSF CCF-2112665 (TILOS), NSF 1730158 CI-New: Cognitive Hardware and Software Ecosystem Community Infrastructure (CHASE-CI), NSF ACI-1541349 CC*DNI Pacific Research Platform, and gifts from Qualcomm.\\
\indent Digital Object Identifier (DOI): see top of this page.
}


\end{abstract}


%
\IEEEpeerreviewmaketitle

\section{Introduction}
\IEEEPARstart{M}{ost} solutions for robotic manipulation today rely on highly structured setups that allow for full state information, fine-grained robot calibration, and predefined action sequences. This requires substantial engineering effort, and the resulting system is intolerant to changes in the environment. Visual feedback from a mounted camera has gained popularity as an inexpensive tool for relaxing the assumption of full state information \cite{Lenz2015DeepLF, Levine2018LearningHC, Xiang2018PoseCNNAC, Zeng2019TossingBotLT}, and Reinforcement Learning (RL) has emerged as a promising technique for learning flexible control policies without the need for detailed human engineering \cite{Williams2004SimpleSG,Sutton1999PolicyGM, Lillicrap2016ContinuousCW, haarnoja2018soft}. Together, vision-based RL could enable use of robots in unstructured environments through flexible policies operating directly from visual feedback, without assuming access to any state information \cite{levine2016end, Pinto2016SupersizingSL, Nair2018VisualRL, Zhan2020AFF}.

However, solving complex precision-based manipulation tasks in an end-to-end fashion remains very challenging, and current methods often rely on important state information that is difficult to obtain in the real world \cite{Gu2017DeepRL,Levine2018LearningHC,Johannink2019ResidualRL,Andrychowicz2020LearningDI,OpenAI2021AsymmetricSF}, and/or rely on known camera intrinsics and meticulous calibration \cite{Xiang2018PoseCNNAC,Shi2020FastUQ} in order to transfer from simulation to the real world. We argue that for a system to be truly robust, it should be able to operate from visual feedback, succeed without the need for camera calibration, and be flexible enough to tolerate task variations, much like humans. In particular, Land et al.~\cite{vision&control} find that when humans perform a complex motor task, the oculo-motor system keeps the centre of gaze very close to the point at which information is extracted, also known as \textit{visual fixation}. However, previous work in visual RL often limit themselves to a single, static third-person monocular camera, which makes extraction of local information from areas of interest challenging, and uncalibrated cameras only exacerbate the problem.

In this work, we propose a setting for vision-based robotic manipulation in which the agent receives visual feedback from two complementary views: (i) a third-person view (\textit{global} information), and (ii) an egocentric view (\textit{local} information), as shown in Figure \ref{fig:teaser}. While the third-person view is static, the egocentric camera moves with the robot gripper, providing the robot with \textit{active} vision capabilities analogous to the oculo-motor system in humans. Because the agent has control over its egocentric vision, it can learn to actively position its camera such that it provides additional information at regions of interest, e.g. accurate localization of points of contact in fine-grained manipulation tasks.

To fuse visual feedback from the two views effectively, we propose to integrate an explicit modeling of visual fixation through a network architecture with soft attention mechanisms. Specifically, we propose to use Transformers \cite{Vaswani2017AttentionIA} with a \textit{cross-view} attention module that explicitly models spatial attention from one view to another. Each view is encoded using separate ConvNet encoders, and we fuse their corresponding feature maps by applying our cross-view attention module bidirectionally. In this way, we let every spatial region in the egocentric view \emph{highlight} the corresponding regions of interest in the third-person view, and vice-versa. We use the learned features as input for an RL policy and optimize the network end-to-end using a reward signal. This encourages the agent to actively control the egocentric camera in a way that maximizes success.

\begin{figure}
    \centering
    \includegraphics[width=0.42\textwidth]{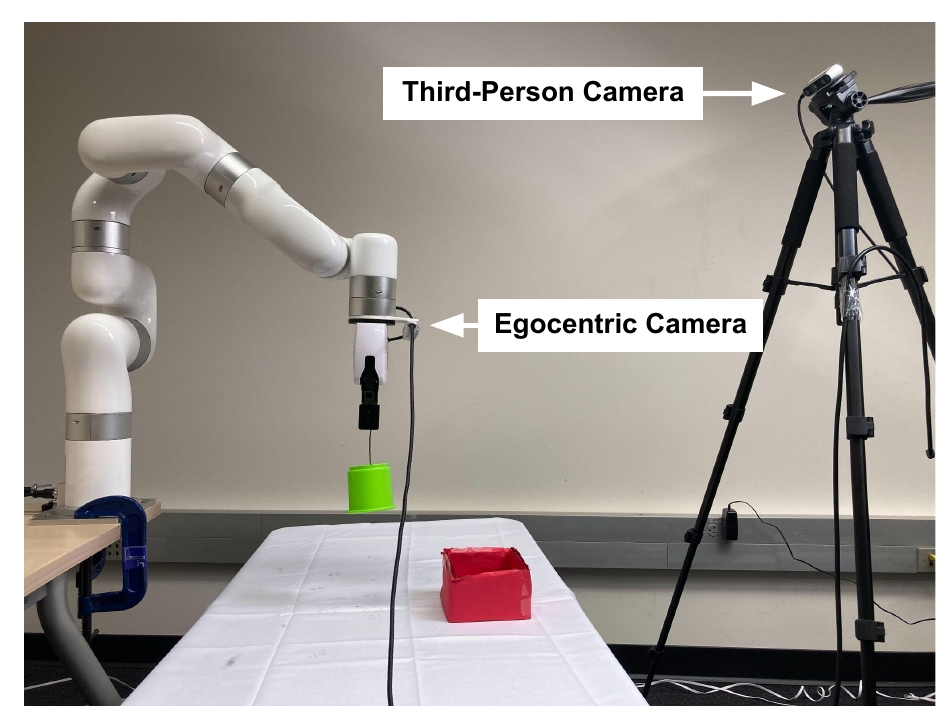}
    \vspace{-0.05in}
    \caption{\textbf{Real robot setup.} Image depicting our real-world environment for the \textit{Peg in Box} task. The third-person camera is static, and the egocentric camera moves along with the robot arm. See Figure \ref{fig:teaser} for camera view samples.}
    \label{fig:real_setup}
    \vspace{-0.25in}
\end{figure}

To demonstrate the effectiveness of our method, we conduct extensive empirical evaluation and compare to a set of strong baselines on four precision-based manipulation tasks: (1) \textit{Reach}, a task in which the robot reaches for a goal marked by a red disc placed on the table,
(2) \textit{Push}, a task in which the robot pushes a cube to a goal marked by a red disc, both placed on the table,
(3) \textit{Peg in Box}, a task in which the objective is to insert a peg tied to the robot's end-effector into a box placed on the table,
(4) \textit{Hammer}, a task in which the objective is to hammer in an out-of-position peg.
Each of the four tasks are shown in Figure \ref{fig:teaser}. We find that our multi-view setting and cross-view attention modules each improve learning over single-view baselines, and we further show that our method successfully transfers from simulation to a real robot setup (shown in Figure \ref{fig:real_setup}) with uncalibrated cameras, no access to state information, and a high degree of task configuration variability. In the challenging \textit{Hammer} task, our method is significantly more successful during transfer than any of our baselines, achieving a success rate of $\mathbf{75\%}$ versus $38\%$ for a multi-view baseline without cross-view attention, and only $13\%$ and $0.0\%$ for our single-view baselines. Finally, we qualitatively observe that (i) multi-view methods appear less prone to error due to lack of camera calibration, and (ii) our proposed cross-view attention mechanism improves precision in tasks that require fine-grained motor control.

%
\vspace{-0.025in}
\section{RELATED WORK}
\vspace{-0.025in}

\textbf{Vision-Based Robotic Manipulation.} Learning RL policies for end-to-end vision-based robotic manipulation via task supervision has been widely explored \cite{levine2016end,Lillicrap2016ContinuousCW,Pinto2016SupersizingSL,Nair2018VisualRL,haarnoja2018soft, Kostrikov2021ImageAI, Yan2020LearningPR, Zhan2020AFF}. For example, Lillicrap et al.~\cite{Lillicrap2016ContinuousCW} solve simulated continuous control tasks directly from images with no access to state information, and Zhan et al.~\cite{Zhan2020AFF} shows that vision-based RL can solve real-world manipulation tasks efficiently given a small number of expert demonstrations. Despite this progress it is still very challenging -- especially for vision-based policies -- to transfer learned skills to other environments, e.g. Sim2Real \cite{Pinto2018AsymmetricAC,Cobbe2019QuantifyingGI,Hansen2021SelfSupervisedPA}. To facilitate transfer, related work also leverage important state information such as object pose, which is readily available in simulation but difficult to obtain in the real world \cite{Gu2017DeepRL,Levine2018LearningHC,Johannink2019ResidualRL,Andrychowicz2020LearningDI,Shi2020FastUQ,OpenAI2021AsymmetricSF}. While this approach is valid, it is limited by the accuracy of pose estimation algorithms which typically require a well calibrated system and/or reference models \cite{Xiang2018PoseCNNAC,Li2018AUF,Shi2020FastUQ,Wang2019DenseFusion6O}. In comparison, our approach learns directly from raw images and transfers to a real robot with uncalibrated cameras.

\textbf{Multi-View Robotic Manipulation.} Multi-modal sensor fusion is a well-studied problem in robotics \cite{levine2016end, Lee2019MakingSO, OpenAI2021AsymmetricSF, Yang2021LearningVQ}. For example, Lee et al.~\cite{Lee2019MakingSO} propose a method for fusing vision and haptic feedback without explicit supervision. However, using multiple \textit{views} of the \textit{same} modality remains a relatively underexplored topic that has only recently gained interest \cite{akinola2020learning, OpenAI2021AsymmetricSF, Chen2021UnsupervisedLO}. Akinola et al. \cite{akinola2020learning} show that third-person views from multiple, statically placed cameras improves policy learning in precision-based manipulation tasks over a single view. Specifically, their algorithm aggregates features encoded from multiple cameras by simple addition and performs RL on top of these features. In addition to camera inputs, they assume access to state information from the gripper. OpenAI et al.~\cite{OpenAI2021AsymmetricSF} explore a multi-view robotic manipulation setting similar to ours, but they require privileged state information that is not easily available in the real world, e.g. full object and goal states. Concurrent to our work, Chen et al. \cite{Chen2021UnsupervisedLO} propose a framework for learning 3D keypoints for control from multiple third-person views. While they similarly to Akinola et al. \cite{akinola2020learning} demonstrate benefits from additional views, keypoint detection remains prone to error from uncalibrated cameras. All of the aforementioned works only consider manipulation tasks \textit{in simulation}, and do \textit{not} consider the problem setting of robotic manipulation from egocentric and third-person cameras without access to state information. In this work, we develop a real robotic system that operates strictly from uncalibrated egocentric and third-person cameras.

\textbf{Active Vision for Manipulation.}
Robots operating only from egocentric images suffer from incomplete state information (occlusion, small field-of-view), while when operating only from third-person vision fail to capture more accurate interactions. A natural solution to this problem is active vision. An active vision system is one that can manipulate the viewpoint of the camera(s) in order to investigate the environment and get better information from it \cite{bajcsy1988active, wilkes1994active}. Traditional methods for active vision \cite{chen2011active} either use hand-crafted utility functions \cite{kriegel2015efficient} or are uncertainty or reconstruction based \cite{fraundorfer2012vision}, whereas learning based \cite{hepp2018learn} approaches explicitly learn these utility functions for selecting the next-best view using ground truth labels. In this work, we use the task-based reward to guide active vision. Cheng et al.~\cite{cheng2018reinforcement} propose an active vision system that learns camera control policies directly from task-based rewards, in coordination to the manipulation policies. However, they use a separate camera to actively focus on objects of interest which requires an additional manipulator. In a similar fashion, Zaky et al. \cite{Zaky2020ActivePA} use an additional manipulator to achieve an active perception system. Importantly, these approaches do not show real-world results. Our approach employs an egocentric camera tied to the wrist of the robot for active vision. While it is hard to both carry out manipulation and keep objects of interest in view in this perspective, our two complementary cameras enable us to achieve this through cross-view attention.

\textbf{Attention Mechanisms in RL.} Our multi-view fusion builds upon the soft attention mechanism which has been widely adopted in natural language processing \cite{Vaswani2017AttentionIA,Devlin2019BERTPO} and computer vision \cite{Wang2018NonlocalNN,Dosovitskiy2021AnII}, and has recently been popularized in the context of RL \cite{Jiang2018LearningAC,Chen2021DecisionTR,Janner2021ReinforcementLA,Hansen2021StabilizingDQ,Yang2021LearningVQ, James2021CoarsetoFineQE}. For example, Jiang et al. \cite{Jiang2018LearningAC} propose an attention-based communication model that scales to cooperative decision-making between a large amount of agents, and Yang et al. \cite{Yang2021LearningVQ} use a cross-modal Transformer to fuse state information with a depth input in simulated locomotion tasks. James et al. \cite{James2021CoarsetoFineQE} is a parallel work that is closest to our approach in their idea of using visual attention to attend to parts of an input RGB image in robotic manipulation tasks. However, like most other works on image-based RL, they restrict their study to a single, fixed third-person view. We propose a \textit{cross-view} attention mechanism that fuses information from multiple camera views, and we are -- to the best of our knowledge -- the first to demonstrate the effectiveness of attention-based policies in transfer from simulation to a real robot setup.

\section{BACKGROUND}
\label{sec:background}
\textbf{Vision-Based Reinforcement Learning.}
We formulate interaction between the agent's vision-based control policy and environment as an infinite-horizon Partially Observable Markov Decision Process (POMDP) \cite{kaelbling1998pomdp} described by a tuple $(\mathcal{S}, \mathcal{A}, \mathcal{P}, p_{0}, r, \gamma)$, where $\mathcal{S}$ is the state space, $\mathcal{A}$ is the action space, $\mathcal{P}\colon \mathcal{S} \times \mathcal{A} \mapsto \mathcal{S}$ is a transition function that defines a conditional probability distribution $\mathcal{P}(\cdot|\mathbf{s}_{t},\mathbf{a}_{t})$ over possible next states given current state $\mathbf{s}_{t} \in \mathcal{S}$ and action $\mathbf{a}_{t} \in \mathcal{A}$ taken at time $t$, $P_{0}$ is a probability distribution over initial states, $r \colon \mathcal{S} \times \mathcal{A} \mapsto \mathbb{R}$ is a scalar reward function, and $\gamma \in [0, 1)$ is a discount factor. In a POMDP, the underlying state $\mathbf{s}$ of the system is assumed unavailable, and the agent must therefore learn to implicitly model states from raw image observations. Our goal is to learn a policy $\pi \colon \mathcal{S} \mapsto \mathcal{A}$ that maximizes return $\mathbb{E}_{\Gamma\sim\pi}[\sum_{t=0}^{\infty} \gamma^{t} r(\mathbf{s}_{t}, \mathbf{a}_{t})]$ along a trajectory $\Gamma = (\mathbf{s}_{0},\mathbf{s}_{1},\dots,\mathbf{s}_{\infty})$ where $\mathbf{s}_{0}\sim P_{0}$, $\mathbf{a}_{t} \sim \pi(\cdot | \mathbf{s}_{t})$, and $\mathbf{s}_{t+1} \sim \mathcal{P}(\cdot | \mathbf{s}_{t}, \mathbf{a}_{t})$, and we use the Soft Actor-Critic \cite{haarnoja2018soft} RL algorithm for policy search.

\begin{figure}
    \centering
    \includegraphics[width=0.45\textwidth]{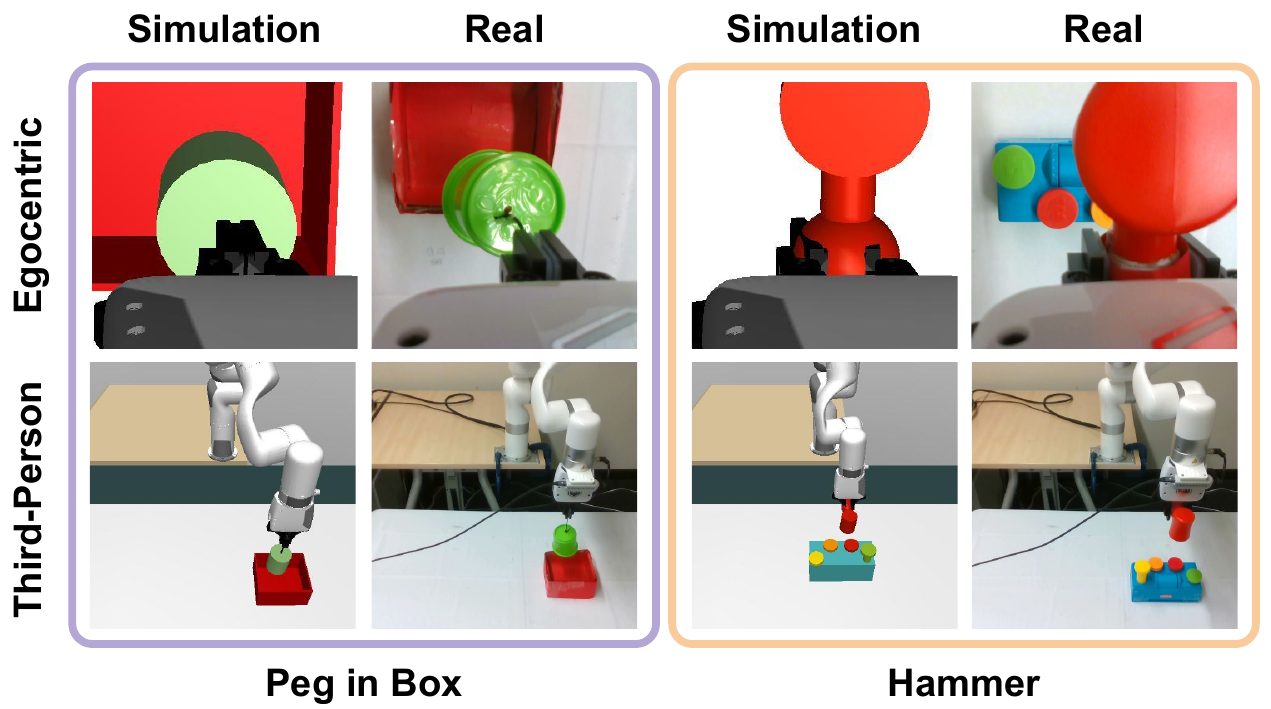}
    \vspace{-0.025in}
    \caption{\textbf{Sim2Real.} Samples from our simulation and real world environments for each view and two tasks, \textit{Peg in Box} and \textit{Hammer}. We emphasize that the real world differs in both visuals, dimensions, dynamics, and camera views, but roughly implement the same task as in simulation. Also note that there is only an approximate correspondence between the samples from simulation and the real world shown here.}
    \label{fig:camera-views}
    \vspace{-0.2in}
\end{figure}

\textbf{Soft Actor-Critic}
(SAC) \cite{haarnoja2018soft} is an off-policy actor-critic algorithm that learns a (soft) state-action value function $Q_{\theta} \colon \mathcal{S} \times \mathcal{A} \mapsto \mathbb{R}$, a stochastic policy $\pi_{\theta}$ as previously defined, and optionally a temperature parameter $\tau_{\theta}$. $Q_{\theta}$ is optimized to minimize the (soft) Bellman residual \cite{Sutton1988LearningTP}, $\pi_{\theta}$ is optimized using a maximum entropy objective \cite{Bagnell2010ModelingPA,Ziebart2008MaximumEI}, and $\tau_{\theta}$ is optimized to maintain a desired expected entropy; see \cite{haarnoja2018soft} for further details. For brevity, we generically refer to learnable parameterization by a subscript $\theta$ throughout this work.

\begin{figure*}[t]
    \centering
    \includegraphics[width=0.94\textwidth]{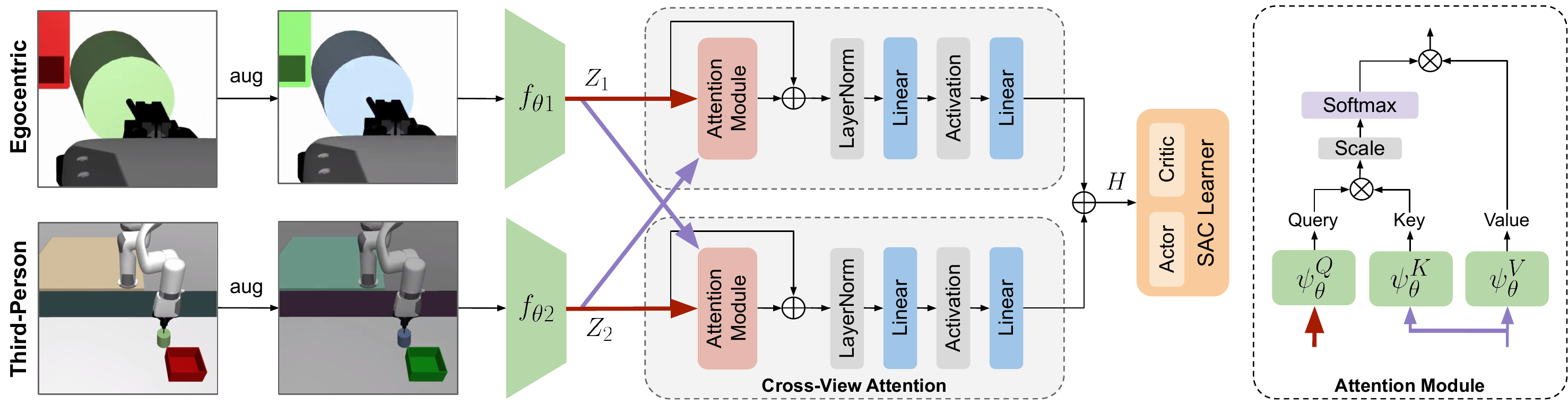}
    \vspace{-0.05in}
    \caption{\textbf{Architectural overview.} Egocentric and third-person views $O_{1},O_{2}$ are augmented using stochastic data augmentation, and are encoded using separate ConvNet encoders to produce spatial feature maps $Z_{1}, Z_{2}$. We perform cross-view attention between views using a Transformer such that features in $Z_{1}$ are used as queries for spatial information in $Z_{2}$, and vice-versa. Features are then aggregated using simple addition ($\oplus$ in the figure), and used as input for a Soft Actor-Critic (SAC) policy.}
    \label{fig:architectural-overview}
    \vspace{-0.15in}
\end{figure*}

\textbf{Robot Setup.} Our real-world setup is shown in Figure \ref{fig:real_setup}. We use a Ufactory xArm 7 robot with an xArm Gripper as the robot platform in both our real-world and simulation experiments, although our method is agnostic to the specific robot hardware. The camera poses and the shape and size of objects in the simulation roughly mimic that of the real-world. We use Intel RealSense cameras for our experiments and apply the same pre-processing to both the simulated and real images. Image samples from the simulation and real-world robot setup can be seen in Figure \ref{fig:camera-views}. Using the Mujoco \cite{todorov2012mujoco} simulation engine, we create a simulation environment with the xArm robot and other objects as necessary in the robotic tasks. Two cameras are mounted to overlook the robot workspace where the task is being performed. With respect to the robot, one camera is directly placed in front of the whole setup (third-person view) with a large field of view covering the robot and the workspace, and the other camera (egocentric view) is attached to the robot at its wrist, i.e., right above where the gripper meets the robot. While the third-person camera is fixed, the egocentric camera moves along with the robot and provides a top-down view of the robot workspace. We emphasize that the cameras are \textit{uncalibrated}, that the policy operates strictly from RGB images ($84\times84$ pixels) captured by the two cameras, and that the policy has \textit{no} access to state information.

\section{METHOD}
Inspired by visual fixation in the human eye, we propose to fuse egocentric and third-person views in a multi-camera robotic manipulation setting, leveraging Transformers for explicit modeling of fixation through its attention mechanism. Our method is a general framework for multi-view robotic manipulation that is simple to both implement and deploy in a real-world setting. For simplicity, we describe our method using SAC \cite{haarnoja2018soft} as backbone learning algorithm as used in our experiments, but we emphasize that our method is agnostic to the particular choice of learning algorithm. We introduce each component of our method in the following.

\subsection{Multi-View Robotic Manipulation}
We propose to learn vision-based RL policies for robotic manipulation tasks using raw, uncalibrated RGB inputs from two complementary camera views: (i) an egocentric view $O_{1}$ from a camera mounted on the wrist of the robot, and (ii) a fixed third-person view $O_{2}$. The third-person view provides global information about the scene and robot-object relative configurations, while the egocentric view serves to provide local information for fine-grained manipulation; see Figure \ref{fig:camera-views} for samples. Because the egocentric camera is attached to the robot, it provides the robot with \textit{active} vision capabilities -- control over one of the two views. Our method learns a joint latent representation from the two camera views, encouraging the model to relate global and local information, and actively position itself such that the egocentric view provides valuable local information for fine-grained manipulation.

\subsection{Network Architecture}
Figure \ref{fig:architectural-overview} provides an overview of our method and architecture. Our proposed architecture takes two raw RGB images $O_{1}, O_{2} \in \mathbb{R}^{C\times H\times W}$ from the egocentric and third-person cameras, respectively (s.t. the system state $\mathbf{s}$ is approximated by $\{O_{1}, O_{2}\}$), and encodes them separately using ConvNet encoders $f_{\theta 1}$ and $f_{\theta 2}$ to produce latent feature maps $Z_{1}, Z_{2} \in \mathbb{R}^{C'\times H'\times W'}$. We propose to fuse the two feature maps using a Transformer \cite{Vaswani2017AttentionIA} with a cross-view attention module that takes $Z_{1}, Z_{2}$ as input and produces a single feature map $H = T_{\theta}(Z_{1}, Z_{2})$. The aggregated features $H$ are flattened and fed into the policy (actor) $\pi_{\theta}$ and state-action value function (critic) $Q_{\theta}$ s.t. $\mathbf{a} \sim \pi_{\theta}(\cdot | T_{\theta}(f_{\theta 1}(O_{1}), f_{\theta 2}(O_{2})))$ and similarly for the critic. In practice, we optimize all components end-to-end using the actor and critic losses of SAC, but only back-propagate gradients from the critic to encoder components $T_{\theta}, f_{\theta 1}, f_{\theta 2}$, as is common practice in image-based RL \cite{yarats2019improving, srinivas2020curl, hansen2021softda, Kostrikov2021ImageAI}. In the following section, we describe our cross-view attention mechanism in more detail. 

\begin{figure}[t!]
    \centering
    \includegraphics[width=0.4\textwidth]{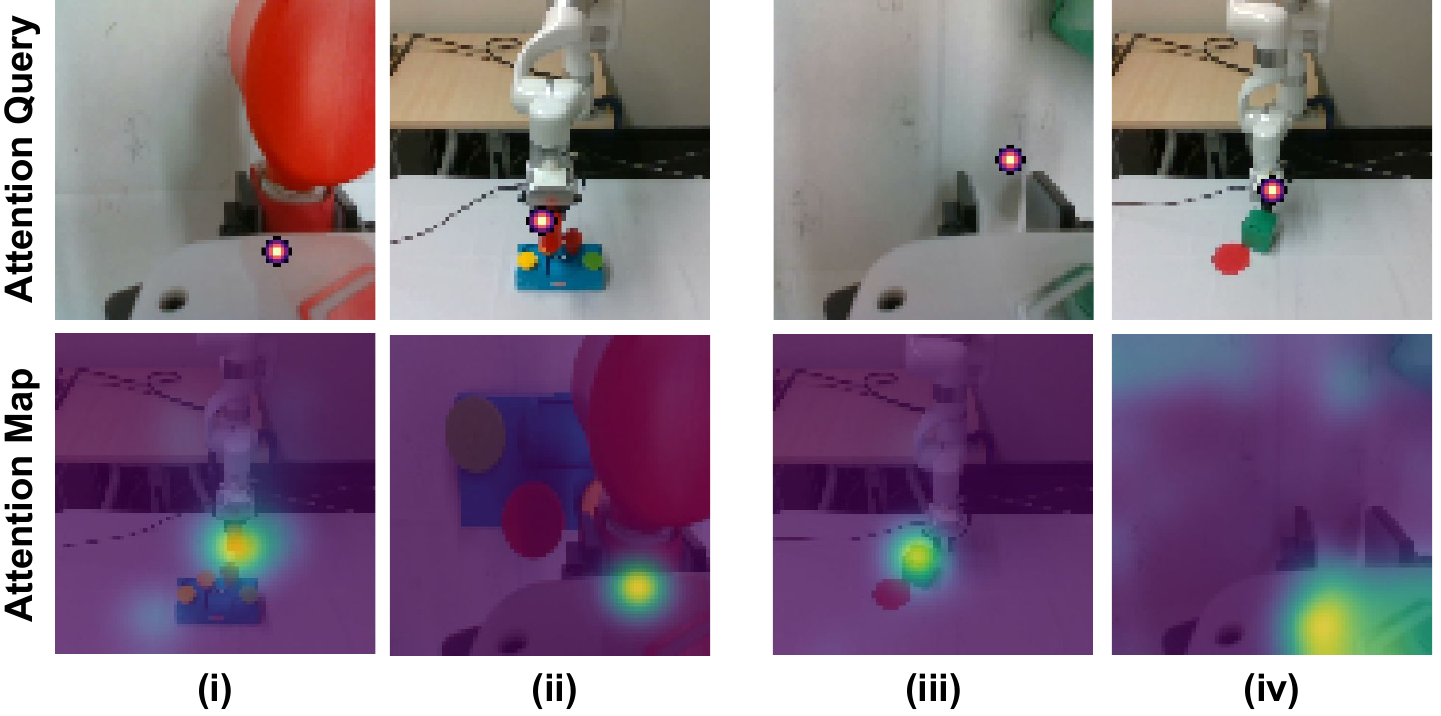}
    \vspace{-0.05in}
    \caption{\textbf{Attention maps.} Visualization of attention maps learned by our policy for \textbf{(i), (ii)} \textit{Hammer} task, and \textbf{(iii), (iv)} \textit{Push} task. For samples \textbf{(i), (iii)} a spatial location in egocentric view (red highlight; top) is used as query and its corresponding attention map for the third-person image (green highlight; bottom) is shown. For samples \textbf{(ii), (iv)} the third-person view is used as query and attention maps for the egocentric view are shown. 
    }
    \label{fig:attn_map}
    \vspace{-0.15in}
\end{figure}

\subsection{Cross-View Attention}
\label{sec:cross-view-attention}
Our Transformer $T_{\theta}$ learns to perform cross-view attention by associating spatial information between intermediate feature maps $Z_{1},Z_{2}$ from the egocentric view $O_{1}$ and third-person view $O_{2}$, respectively. $T_{\theta}$ takes as input $Z_{1},Z_{2}$ and produces a joint embedding
\begin{align}
\label{eq:cross-view-attention}
H & = T_{\theta}(Z_{1},Z_{2})\\
\label{eq:cross-view-attention-both-dir}
& \triangleq g_{\theta1}\left(\operatorname{LN}(Z_{1} + V_{2} A_{12}) \right) + g_{\theta2}\left(\operatorname{LN}(Z_{2} + V_{1} A_{21}) \right)
\end{align}
where $g_{\theta1},g_{\theta2}$ are Multi-Layer Perceptrons (MLP), $\operatorname{LN}$ is a $\operatorname{LayerNorm}$ \cite{Ba2016LayerN} normalization, and $A_{12},A_{21}$ are normalized (soft) scaled dot-product attention \cite{Vaswani2017AttentionIA} weights
\begin{align}
    \label{eq:self-attention}
    & A_{12} = \sigma\left( K_{2}^{\top} Q_{1} \big/ \sqrt{C'} \right),~A_{21} = \sigma\left( K_{1}^{\top} Q_{2} \big/ \sqrt{C'} \right)
\end{align}
for $Q_{i}, K_{j}, V_{j} \in \mathbb{R}^{C' \times H' \times W'}$ denoted as the queries, keys, and values, respectively, for the cross-view attention between $Z_{i}$ and $Z_{j}$, and $\sigma$ is a $\operatorname{Softmax}$ normalization. $Q_{i}, K_{j}, V_{j}$ are view-dependent embeddings
\begin{align}
    \label{eq:qvk}
    & Q_{i} = \psi^{Q}_{\theta i}(\operatorname{LN}(Z_{i})),~K_{j} = \psi^{K}_{\theta j}(\operatorname{LN}(Z_{j}))\\
    & V_{j} = \psi^{V}_{\theta j}(\operatorname{LN}(Z_{j})) \,,
\end{align}
where $\psi^{Q}_{\theta i},\psi^{K}_{\theta j},\psi^{V}_{\theta j}$ are $1\times1$ convolutional layers.

Intuitively, each of the two cross-view attention mechanisms between feature maps $Z_{i},Z_{j}$ can be interpreted as a differential (spatial) lookup operation, where $Z_{i}$ is used as query to retrieve information in $Z_{j}$. By performing cross-view attention bidirectionally as shown in Equation \ref{eq:cross-view-attention-both-dir}, $T_{\theta}$ enables flow of spatial information both from the egocentric view to the third-person view and vice-versa. With spatial attention, the policy can learn to associate concepts present in both views, e.g. objects or the gripper as shown in Figure \ref{fig:attn_map}, and enhance both object-centric features as well as the overall 3D geometry of the scene. We note that, while we consider two views in this work, our method can trivially be extended to $n$ views by letting $T_{\theta}$ compute cross-view attention bidirectionally between all pairs of views.

\subsection{Data Augmentation}
To aid both learning in simulation and transfer to the real world, we apply data augmentation to image observations, both in our method and across all baselines. During training, we sequentially apply stochastic image shifts of 0-4 pixels as in Kostrikov et al. \cite{Kostrikov2021ImageAI}, as well as the color jitter augmentation from Hansen et al. \cite{Hansen2021StabilizingDQ}. Augmentations are applied independently to each view, i.e., $O^{\mathrm{aug}}_{1} = \operatorname{aug}(O_{1}, \zeta_{1}),~O^{\mathrm{aug}}_{2} = \operatorname{aug}(O_{2}, \zeta_{2}),~\zeta_{1},\zeta_{2} \sim \Omega$ where $\zeta_{1},\zeta_{2}$ parameterize the augmentations and $\Omega$ is a uniform distribution over the joint augmentation space. 

\section{EXPERIMENTS}
We investigate the effects of our proposed multi-view setting as well as our Transformer's cross-view attention mechanism on a set of precision-based robotic manipulation tasks from visual feedback. Both our method and baselines are trained entirely in simulation using dense rewards and randomized initial configurations of the robot, goal, and objects across the workspace. We evaluate methods both in the simulation used for training, and a real robot setup as described in Section \ref{sec:background}. For consistent results, we report success rate over a set of pre-defined goal and object locations both in simulation and the real world. Goal locations vary between tasks, and the robot is reset after each trial. 

\textbf{Tasks.} Figure \ref{fig:teaser} provides samples for each task and view considered. Using the simulation setup as described in Section \ref{sec:background}, we consider the following tasks: \textbf{(1) Reach}, a task in which the robot reaches for a goal marked by a red disc placed on the table. Success is considered when the robot gripper reaches within 5 cm of the goal.
\textbf{(2) Push}, a task in which the robot pushes a cube to a goal marked by a red disc, both placed on the table. Success is considered when the cube reaches within 10 cm of the goal.
\textbf{(3) Peg in Box}, a task in which the objective is to insert a peg tied to the robot's end-effector into a box placed on the table. Success is considered when the peg reaches within 5 cm of the goal and as a result gets inserted into the box.
\textbf{(4) Hammer}, a task in which the objective is to hammer in an out-of-position peg. At each episode, one peg is randomly selected from 4 differently colored pegs. Success is considered when the out-of-position peg is hammered back within 1 cm of the box.
Both the \textit{Reach} and \textit{Push} tasks use an XY action space, with movement along Z constrained. \textit{Peg in Box} and \textit{Hammer} use an XYZ action space. Episodes are terminated when the policy succeeds, collides with the table, or a maximum number of time steps is reached. An episode is counted as a success if the success criteria defined above is met for any time step in the episode.

\textbf{Setup.} We implement our method and baselines using Soft Actor-Critic (SAC) \cite{haarnoja2018soft} as learning algorithm and, whenever applicable, we use the same network architecture and hyperparameters as in previous work on image-based RL \cite{hansen2021softda, Hansen2021StabilizingDQ}, i.e., $f_{\theta}$ consists of $11$ convolutional layers. Observations are RGB images of size $84\times84$ from either one or two cameras depending on the method. Although related works commonly approximate the system state $\mathbf{s}$ using a stack of frames, we empirically find a single frame sufficient for both learning and transfer. All methods are trained for $500$k frames and evaluated for $30$ trials across 5 object locations, both in simulation and on the real robot. However, for the \textit{Hammer} task, we conduct $24$ trials: $2$ trials for each of the $4$ pegs, repeated across $3$ object locations, $2\times4\times3=24$. In simulation, all results are averaged over $3$ model seeds; in real world experiments, we report transfer results for the best seed of each method due to real world constraints.

\begin{figure*}[t!]
    \centering
    \includegraphics[width=0.84\textwidth]{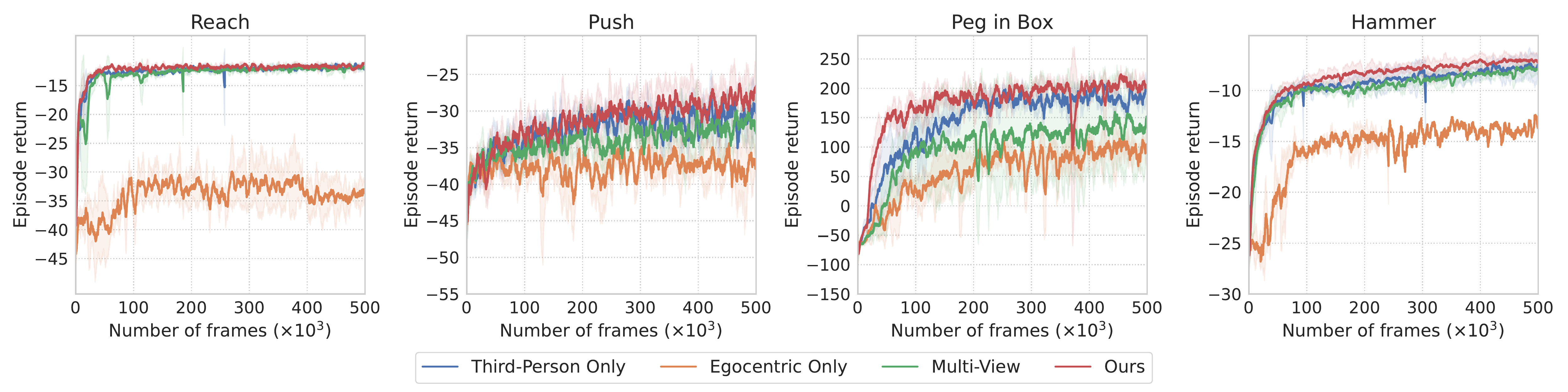}
    \vspace{-0.075in}
    \caption{\textbf{Training performance.} Episode return as a function of the number of frames during training, averaged over $3$ seeds and shaded area is $\pm1$ std. deviation. Our method consistently performs on par or better than all baselines considered.}
    \label{fig:learning-curves}
    \vspace{-0.175in}
\end{figure*}

\textbf{Baselines.} We compare our method to a set of strong baselines, all using Soft Actor-Critic (SAC) as learning algorithm and the same choice of network architecture, hyperparameters, and data augmentations as our method. Specifically, we compare our method to the following baselines: \textbf{(1) Third-Person View} (\textit{3rd}) using visual feedback from a fixed third-person camera, \textbf{(2) Egocentric View} (\textit{Ego}) using only the egocentric camera mounted on the robot's wrist, \textbf{(3) Multi-View} (\textit{Multi}) that uses both views as input, encodes each view using separate encoders, and aggregates extracted features using addition. We emphasize that all baselines are implemented using the same image augmentations (image shift and color jitter) as our method, which makes them largely equivalent to DrQ \cite{Kostrikov2021ImageAI}, a state-of-the-art algorithm for image-based RL that builds on SAC. Lastly, we note that (3) implements our method without cross-view attention and is therefore analogous to the method proposed by Akinola et al. \cite{akinola2020learning} but for a combination of third-person and egocentric views, without state information from the gripper, and including recent advances in image-based RL such as augmentations. In addition to these baselines, we also ablate our design choices.

\begin{table}[t!]
\caption{\textbf{Simulation experiments.} Success rate of our method and baselines when trained and evaluated in simulation. Averaged across $3$ seeds and $30$ trials ($24$ for \textit{Hammer}). Our method is significantly more success in \textit{Hammer}.}
\label{tab:sim-experiments}
\centering
\resizebox{0.36\textwidth}{!}{%
\begin{tabular}{lcccc}
\toprule
\texttt{Simulation}     & 3rd   & Ego  & Multi   & Ours \\ \midrule
Reach                   & $\mathbf{1.00}$ & $0.15$ & $\mathbf{1.00}$ & $\mathbf{1.00}$ \vspace{0.015in} \\ 
Push                    & $0.75$ & $0.50$ & $\mathbf{0.80}$ & $\mathbf{0.80}$ \vspace{0.015in} \\ 
Peg in Box              & $\mathbf{0.80}$ & $0.20$ & $\mathbf{0.80}$ & $\mathbf{0.80}$ \vspace{0.015in} \\
Hammer                  & $0.30$ & $0.04$ & $0.50$ & $\mathbf{0.86}$ \vspace{0.015in} \\ \bottomrule
\end{tabular}
}
\vspace{-0.05in}
\end{table}

\begin{table}[t!]
\caption{\textbf{Real robot experiments.} Success rate of our methods when trained in simulation and transferred to a real robot. Averaged across $30$ trials ($24$ for \textit{Hammer}).}
\label{tab:real-robot-experiments}
\centering
\resizebox{0.36\textwidth}{!}{%
\begin{tabular}{lcccc}
\toprule
\texttt{Real robot}     & 3rd    & Ego  & Multi & Ours \\ \midrule
Reach                   & $0.83$ & $0.17$ & $0.83$ & $\mathbf{1.00}$ \vspace{0.015in} \\ 
Push                    & $0.10$ & $0.17$ & $0.23$ & $\mathbf{0.80}$ \vspace{0.015in} \\ 
Peg in Box              & $0.23$ & $0.27$ & $0.50$ & $\mathbf{0.80}$ \vspace{0.015in} \\
Hammer                  & $0.13$ & $0.00$ & $0.38$ & $\mathbf{0.75}$ \vspace{0.015in} \\ \bottomrule
\end{tabular}
}
\vspace{-0.2in}
\end{table}

\subsection{Robotic Manipulation in Simulation}
\label{sec:simulations}
\vspace{-0.025in}
Before discussing our real robot experiments, we first consider methods in simulation. Training performances are shown in Figure \ref{fig:learning-curves}. Our method using both multi-view inputs and a Transformer performs on par or better than baselines on all tasks, and we particularly observe improvements in sample efficiency on the \textit{Peg in Box} task that requires 3D geometric understanding. All methods except the egocentric-only baselines trivially solve the \textit{Reach} task, but we include it to better ground our results. We conjecture that the egocentric baseline performs significantly worse than other methods due to its lack of global scene information.

\begin{figure}[t!]
    \centering
    \includegraphics[width=0.4\textwidth]{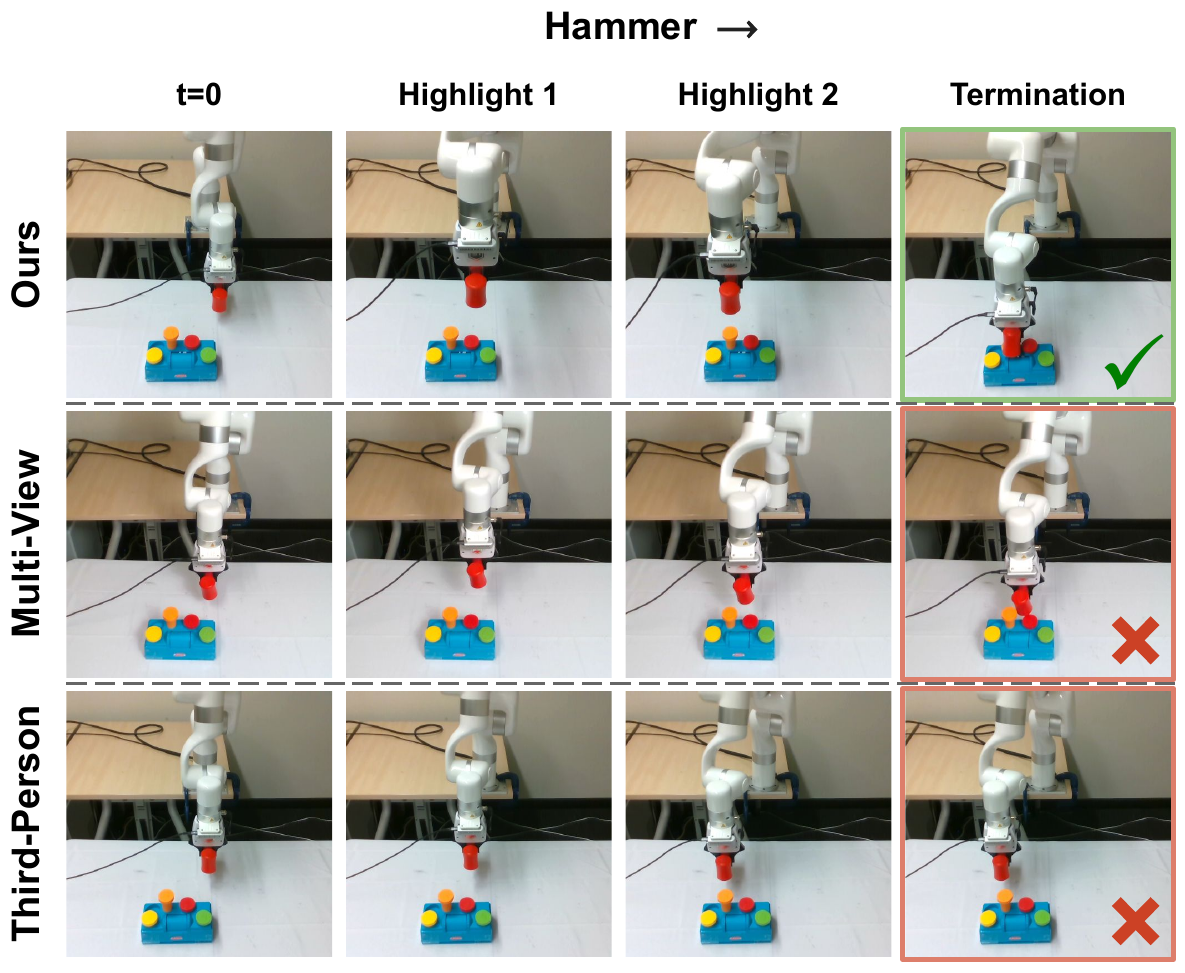}
    \vspace{-0.05in}
    \caption{\textbf{Qualitative results.} Sample trajectories from the \textit{Hammer} task for our method, the multi-view baseline using both camera views, as well as the baseline using only a third-person view. Episodes are terminated when the policy succeeds, collides with the table, or a maximum number of time steps is reached. Our method succeeds in $75\%$ of trials.}
    \label{fig:sample-trajectories}
    \vspace{-0.225in}
\end{figure}

The corresponding success rates for each task and method after training for 500k frames are shown in Table \ref{tab:sim-experiments}. Interestingly, while a third-person view is sufficient for learning to solve $3$ out of $4$ tasks, we observe substantial improvements in success rate in both our method ($\mathbf{+56\%}$) and multi-view ($+20\%)$, which is not immediately obvious from the episode returns. Qualitatively, we find that the third-person baseline often approaches the peg but fails to hit it, whereas our method is comparably better at precise manipulation.

\subsection{Sim2Real Transfer}
\label{sec:real-robot-experiments}
We now consider deployment of the learned policies in a Sim2Real setting, i.e., the policies trained in simulation are transferred to a real robot setup as shown in Figure \ref{fig:camera-views}. Success rates on the real robot are shown in Table \ref{tab:real-robot-experiments}.

We find that success of the single-view baselines drops considerably when transferring to the real world. For example, third-person baseline (\textit{3rd}) achieves success rates of $75\%$ and $80\%$ for the \textit{Push} and \textit{Peg in Box} tasks, respectively, in simulation, while merely $10\%$ and $23\%$ in the real world. We conjecture that this drop in performance is due to the reality gap -- and the lack of camera calibration in particular. With the addition of an egocentric view, the multi-view baseline improves transfer to $23\%$ and $50\%$ on the two tasks. Finally, we observe no drop in success for our method on the same two tasks, achieving a success rate of $80\%$ in both tasks, and only a small drop in success rate on the challenging \textit{Hammer} task that requires 3D understanding and a high level of precision. Specifically, our method succeeds in $75\%$ of trials in the \textit{Hammer} task versus only $38\%$ and $13\%$ for the multi-view and single-view baselines, respectively.

We also study the \textit{qualitative} behavior of policies when transferring to the real world. Figure \ref{fig:sample-trajectories} shows sample trajectories for our method, the multi-view baseline, and the third-person baseline on the \textit{Hammer} task. In the \textit{Hammer} task, we observe that the multi-view baseline frequently misses its target by a small margin (presumably due to the reality gap), and the third-person baseline systematically fails to reach the peg, which we conjecture is due to error in 3D perception from the uncalibrated camera. Using our method, we observe that the robot frequently positions its gripper above the peg, such that it is visible from the egocentric view during the hammering motion. Finally, we also evaluate the qualitative behavior of the cross-view attention module. Attention maps for a set of spatial queries are shown in Figure \ref{fig:attn_map}. We find that the agent often attends to regions of interest, such as objects or the gripper. Transformer-based policies could therefore also be a promising technique for explainability in RL.

\begin{table}[t!]
\caption{\textbf{Ablations.} Success rate of our method and ablations when trained in simulation and evaluated in \textit{(left)} the simulated environment, and \textit{(right)} our real robot setup. $A_{12}$ ablates our method by only performing cross-attention from the egocentric view to the third-person view, and $A_{21}$ corresponds to the opposite direction. Averaged across $3$ seeds and $30$ trials ($24$ trials for \textit{Hammer}).}
\label{tab:ablations}
\centering
\resizebox{0.465\textwidth}{!}{%
\begin{tabular}{lccc|ccc}
& \multicolumn{3}{c}{\texttt{Simulation}} & \multicolumn{3}{c}{\texttt{Real robot}} \\ \midrule
     & $A_{12}$ & $A_{21}$   & Ours & $A_{12}$ & $A_{21}$   & Ours \\ \midrule
Reach                   & $\mathbf{1.00}$ & $\mathbf{1.00}$ & $\mathbf{1.00}$ & $0.63$ & $0.73$ & $\mathbf{1.00}$ \vspace{0.01in} \\ 
Push                    & $0.63$ & $0.65$ & $\mathbf{0.80}$ & $0.26$ & $0.37$ & $\mathbf{0.80}$ \vspace{0.01in}\\ 
Peg in Box              & $0.70$ & $0.70$ & $\mathbf{0.80}$ & $0.23$ & $0.53$ & $\mathbf{0.80}$ \vspace{0.01in}\\
Hammer                  & $0.60$ & $0.50$ & $\mathbf{0.86}$ & $0.42$ & $0.50$ & $\mathbf{0.75}$ \vspace{0.01in} \\ \bottomrule
\end{tabular}
}
\vspace{-0.05in}
\end{table}

\begin{table}[t!]
\caption{\textbf{Robot state} as additional input. Success rate comparison when trained in simulation and evaluated on the real robot, along with the \%-change in performance as compared to image-only observations.}
\label{tab:real-state}
\centering
\resizebox{0.30\textwidth}{!}{%
\begin{tabular}{lccc}
\toprule
\texttt{Real robot}     & 3rd             & Multi                & Ours             \\ \midrule
Peg in Box              & $0.30$ & $0.77$   & $\mathbf{0.87}$ \vspace{0.015in} \\
Hammer                  & $0.16$ & $0.43$   & $\mathbf{0.76}$ \vspace{0.015in} \\ \bottomrule
\end{tabular}
}
\vspace{-0.05in}
\end{table}

\begin{table}[t!]
\caption{\textbf{Camera randomization.} Success rates when trained in simulation and evaluated in a camera-randomized simulation, along with \%-change in performance.}
\label{tab:randomized-sim-experiments-drop}
\centering
\resizebox{0.37\textwidth}{!}{%
\begin{tabular}{lll}
\toprule
\texttt{Simulation}     & \multicolumn{1}{c}{Multi}   & \multicolumn{1}{c}{Ours}     \\ \midrule
Reach                    & $0.85$ ($-15.0\%$)   & $\mathbf{0.98}$ ($-2.0\%$) \vspace{0.015in} \\ 
Push                    & $0.46$ ($-42.5\%$)   & $\mathbf{0.72}$ ($-10.0\%$) \vspace{0.015in} \\ 
Peg in Box              & $0.24$ ($-70.0\%$)   & $\mathbf{0.55}$ ($-31.3\%$) \vspace{0.015in} \\
Hammer                  & $0.18$ ($-64.0\%$)   & $\mathbf{0.47}$ ($-45.3\%$) \vspace{0.015in} \\ \bottomrule
\end{tabular}
}
\vspace{-0.2in}
\end{table}

\vspace{-0.025in}
\subsection{Ablations}
\label{sec:ablations}
\vspace{-0.025in}
\textbf{Attention mechanism.} Our experiments discussed in Section \ref{sec:simulations} and \ref{sec:real-robot-experiments} ablate the choice of camera views. Our method learns cross-view attention between the egocentric view and the third-person view and models each of the directions individually, i.e., an attention map $A_{12}$ is computed for egocentric~$\rightarrow$~third-person direction, and $A_{21}$ is likewise computed for the opposite direction. We ablate each of these two modules such that cross-view attention is only learned unidirectionally; results are shown in Table \ref{tab:ablations}. Our ablations indicate that unidirectional cross-view attention achieves similar success rates to that of the multi-view baselines in simulation, only improving marginally in the \textit{Hammer} task. In our real robot experiments, we observe performance gains in both of our unidirectional ablations over the multi-view baseline, but not consistently. We observe the biggest improvements in \textit{Push} and \textit{Hammer}, where global-local coordination is especially important. However, we find that our proposed formulation succeeds more often than either ablation and produces more consistent gains across tasks. We conjecture that letting \textit{both} views attend to each other improves flow of information, which in turn improves the expressiveness of the learned joint representation.

\textbf{Robot state utilization.} We study the effect of including robot state along with visual inputs in Table \ref{tab:real-state}. Robot state includes end-effector position and gripper state, and is readily available both in simulation and the real robot. Following the architecture as described in \cite{akinola2020learning} for fusing state and visual inputs, we train the top 3 performing methods in simulation and test on the real robot on our two most difficult tasks. Adding robot state information brings performance gains across all methods, while \textit{3rd} and \textit{Multi} benefit comparably more from the added states. However, our method still outperforms all in this setting. We hypothesize that the baselines have difficulty extracting state information from visual inputs, whereas our proposed fusion mechanism alleviates this issue and is therefore less dependent on the provided state information. The remaining gap in performance can be attributed to the ability of each method to extract object state information from visual inputs. 

\textbf{Robustness analysis.} Camera calibration can be a time consuming process, and typically needs to be repeated every time the camera or environment changes. Therefore, it is desirable to develop methods that can operate directly from uncalibrated cameras, and we hypothesize that our proposed method increases robustness to such settings. To test this hypothesis, we further evaluate methods in simulation under random camera perturbations, simulating deployment with a variety of uncalibrated cameras. Perturbations include camera properties such as camera pose and FOV. Results are shown in Table \ref{tab:randomized-sim-experiments-drop}.
We find that our proposed cross-view attention mechanism is more robust to camera perturbations than the multi-view baseline across all tasks, presumably due to better information fusing across views.
\vspace{-0.025in}
\section{CONCLUSION}
\vspace{-0.025in}
Precise robotic manipulation from visual feedback is challenging. Through experiments in both simulation and the real world, we observe that both our multi-view setting as well as our proposed cross-view attention mechanism improves learning and in particular improves transfer to the real world, even in the challenging setting of Sim2Real with uncalibrated cameras, no state information, and a high degree of task variability. Therefore, we believe our proposed problem setting and method is a promising direction for future research in robotic manipulation from visual feedback.

\ifCLASSOPTIONcaptionsoff
  \newpage
\fi

\vspace{-0.025in}
\bibliography{references}
\bibliographystyle{ieee/IEEEtran}

\addtolength{\textheight}{-12cm}   
\end{document}